\documentclass[10pt,twocolumn,letterpaper]{article}
\usepackage{cvpr} 
\usepackage{multirow}
\usepackage{booktabs}
\usepackage{times}
\usepackage{epsfig}
\usepackage{graphicx}
\usepackage{amsmath}
\usepackage{subfig}
\usepackage{amssymb}
\usepackage{subfiles}
\usepackage[toc,page]{appendix}
\usepackage{floatrow}
\pdfoutput=1 

\usepackage[font=small,labelfont=bf,tableposition=top]{caption}
\newfloatcommand{capbtabbox}{figure}[][\FBwidth]

\DeclareCaptionLabelFormat{andtable}{#1~#2  \&  \tablename~\thetable}

\usepackage{multirow, makecell}
\usepackage{rotating}
\cvprfinalcopy %

\def\cvprPaperID{6280} %

\begin{document}

\title{Learning to Remember: \\ A Synaptic Plasticity Driven Framework for Continual Learning} %

\newcommand*{\affaddr}[1]{#1} %
\newcommand*{\affmark}[1][*]{\textsuperscript{#1}}
\newcommand*{\email}[2]{\texttt{#1}}

\author{%
Oleksiy Ostapenko\affmark[$\mathsection$,*], Mihai Puscas\affmark[$\mathparagraph$,*], Tassilo Klein\affmark[*], Patrick J\"ahnichen\affmark[$\mathsection$], Moin Nabi\affmark[*]\\
\affaddr{\small \affmark[$\mathsection$]Humboldt-Universit\"at zu Berlin},
\affaddr{\affmark[$\mathparagraph$]University of Trento},
\affaddr{\affmark[*]SAP ML Research}
}

\maketitle

\begin{abstract}                  
  Models trained in the context of continual learning (CL) should be able to learn from a stream of data over an indefinite period of time. The main challenges herein are: 1) maintaining old knowledge while simultaneously benefiting from it when learning new tasks, and 2) guaranteeing model scalability with a growing amount of data to learn from. In order to tackle these challenges, we introduce Dynamic Generative Memory (DGM) - synaptic plasticity driven framework for continual learning. DGM relies on conditional generative adversarial networks with learnable connection plasticity realized with neural masking. Specifically, we evaluate two variants of neural masking: applied to (i) layer activations and (ii) to connection weights directly. Furthermore, we propose a dynamic network expansion mechanism that ensures sufficient model capacity to accommodate for continually incoming tasks. The amount of added capacity is determined dynamically from the learned binary mask. We evaluate DGM in the continual class-incremental setup on visual classification tasks. %
\end{abstract}

\section{Introduction}
Conventional Deep Neural Networks (DNN) fail to continually learn from a stream of data while maintaining knowledge. Specifically, reusing old knowledge in new contexts poses a severe challenge. Generally, there are several fundamental obstacles on the way to a continually trainable AI system: the problem of forgetting when learning from new data (catastrophic forgetting), lack of model scalability, i.e. the inability to scale up the model's size with a continuously growing amount of training data, and finally inability to transfer knowledge across tasks.

Several recent approaches \cite{EWC, ZenkePG17, RWalk, Aljundi} try to mitigate forgetting in ANNs while simulating synaptic plasticity directly in the task solving network. %
It is noteworthy that these methods topically tackle the task-incremental scenario, i.e. a separate classifier %
is trained to make predictions about each task. This further implies the availability of oracle knowledge of the task label at inference time. %
Such evaluation is often referred to as multi-head evaluation in which the task label is associated with a dedicated output head. Alternatively, other approaches rely on single-head evaluation \cite{iCarl,RWalk}. Here, the model is evaluated on all classes observed during the training, no matter which task they belong to. While single-head evaluation does not require oracle knowledge of the task label, it also does not reduce the output space of the model to the output space of the task. Thus single-head evaluation represents a harder, yet more realistic setup. Single-head evaluation is predominantly used in class-incremental setup, in which every newly introduced data batch contains examples of one to many new classes.

As opposed to the task-incremental situation, models in class-incremental setup typically require the previously learned information to be replayed when learning new categories  %
\cite{iCarl, RWalk, VCL}. %
The simplest way to accomplish this is by retaining and replaying real samples of previously seen categories to the task solver. However, retaining real samples has several intrinsic implications. First, it is very much against the notion of bio-inspired design, as natural brains do not feature the retrieval of information identical to originally exposed impressions \cite{mayford2012synapses}. Second, as pointed out by \cite{iCarlGAN, iCarl} storing raw samples of previous data can violate data privacy and memory restrictions of real-world applications. Such restrictions are particularly relevant for the vision domain with its continuously growing dataset sizes and rigorous privacy constraints.

In this work, we address the ``strict'' class-incremental setup. That is, we demand a classifier to learn from a stream of data with different classes occurring at different times with no access to previously seen data, i.e. no storing of real samples is allowed. Such a scenario is solely addressed by methods relying on generative memory - a generative network is used to memorize previously seen data distributions, samples of which can be replayed to the classifier at any time. %
Several strategies exist to avoid catastrophic forgetting in generative networks. %
The most successful approaches
rely on deep generative replay (DGR) \cite{DGR} - repetitive retraining of the generator on a mix of synthesized samples of previous categories and real samples of new classes. %
In this work we propose Dynamic Generative Memory (DGM) with learnable connection plasticity represented by parameter level attention mechanism. As opposed to DGR, DGM features a single generator that is able to incrementally learn about new tasks without the need to replay previous knowledge.

Another important factor in the continual learning setting is the ability to scale, i.e. to maintain sufficient capacity to accommodate for a continuously growing amount of information. 
Given invariant resource constraints, it is inevitable that with a growing number of tasks to learn, the model capacity is depleted at some point in time. 
This issue is again exacerbated when simulating neural plasticity with parameter level hard attention masking. 
In order to guarantee sufficient capacity and constant expressive power of the underlying DNN, %
we keep the number of "free" parameters (i.e. to which the gradient updates can be freely applied) constant by expanding the network with exactly the number of parameters that were blocked for the previous task. %

Our contribution is twofold: \textbf{(a)} we introduce Deep Generative Memory (DGM) - an adversarially trainable generative network that features neural plasticity through efficient learning of a sparse attention masks for the network weights (DGMw) or layer activations (DGMa); To the best of our knowledge we are the first to introduce weight level masks that are learned simultaneously with the base network; Furthermore, we conduct it in an adversarial context of a generative model; DGM is able to incrementally learn new information during adversarial training \emph{without the need to replay previous knowledge to its generator}. %
\textbf{(b)} We propose an adaptive network expansion mechanism, facilitating resource efficient continual learning.
In this context, we compare the proposed method to the state-of-the-art approaches for continual learning. Finally, we demonstrate that DGMw accommodates for higher efficiency, better parameter re-usability and slower network growth than DGMa.  %

\section{Related Work}
\label{sec:relWork}
\begin{figure*}[!ht]
    \centering            
    \includegraphics[width=1.0\linewidth]{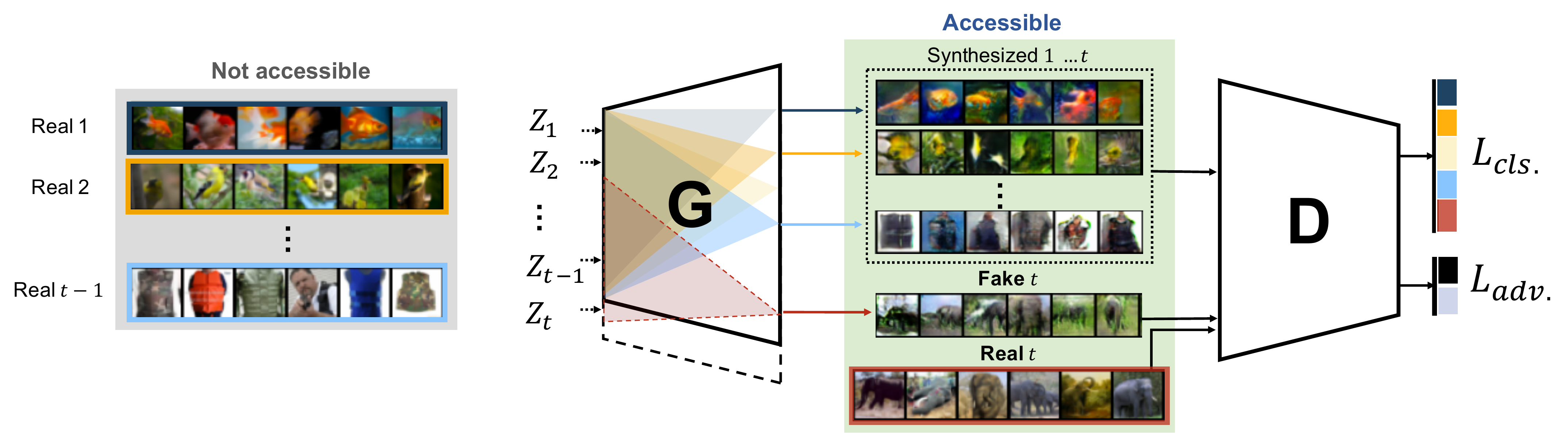}                         
    \caption{Dynamic Generative Memory: auxiliary output of $D$ is trained on the real samples of the current task $t$ and synthesized sample of previously seen tasks $1...t-1$. Adversarial training is accomplished with real and fake samples of the current task. Connection plasticity simulated with binary mask applied to the generators weights or activations is learned simultaneously to the adversarial training.}
    \label{fig1}
\end{figure*}

Among the first works dealing with catastrophic forgetting in the context of lifelong learning are \cite{french1999catastrophic,MCCLOSKEY1989109,Ratcliff90connectionistmodels}, who tackle this problem by employing  shallow neural networks, whereas our method makes use of modern deep architectures. Lately, a wealth of works dealing with catastrophic forgetting in context of DNNs have appeared in the literature, see e.g., \cite{EWC,ZenkePG17, LwF,HAT, Aljundi,rosenfeld2018incremental}. Thus, EWC \cite{EWC} and RWalk \cite{RWalk} rely on Fisher's information to identify parameters that carry most of the information about previously learned tasks, and apply structural regularization to ``discourage'' change of these parameters. \cite{ZenkePG17} and \cite{Aljundi} identify important parameter segments based on the sensitivity of the loss or the learned prediction function to changes in the parameter space. 
Instead of relying on ``soft'' regularization techniques, \cite{HAT} and \cite{RusuRDSKKPH16} propose to dedicate separate parameter subspaces to separate tasks. %
Serr{\`{a}} et al.~\cite{HAT} propose a hard attention to the task (HAT) mechanism. HAT finds dedicated parameter subspaces for all tasks in a single network while allowing them to mutually overlap. The optimal solution is then found in the corresponding parameter subspace of each task. 
All of these methods have been proposed for a ``task-incremental learning'' setup. %
In our work %
we specifically propose a method to overcome catastrophic forgetting within the ``class-incremental'' setup. %
Notably, a method designed for class-incremental learning can be generally applied in a task-incremental setup. %

Several continuous learning approaches \cite{iCarl,VCL,FearNet}, address catastrophic forgetting in the class-incremental setting, i.e. by storing raw samples of previously seen data and making use of them during the training on subsequent tasks. Thus, iCarl \cite{iCarl} proposes to find $m$ most representative samples of each class whose mean feature space most closely approximates the entire feature space of the class. The final classification task is done by the means of the nearest mean-of-exemplars classifier. %

Recently, there has been a growing interest in employing deep generative models for memorizing previously seen data distributions instead of storing old samples. \cite{DGR,2018arXiv180902058W} rely on the idea of generative replay, which requires retraining the generator at each time step on a mixture of synthesized images of previous classes and real samples from currently available data. However, apart from being inefficient in training, these approaches are severely prone to ``semantic drifting''. Namely, the quality of images generated during every memory replay highly depends on the images generated during previous replays, which can result in loss of quality and forgetting over time. In contrast, we propose to utilize a single generator that is able to incrementally learn new information during the normal adversarial training without the need to replay previous knowledge. This is achieved through efficiently learning a sparse mask for the learnable units of the generator network.

Similar to our method, \cite{seff2017continual} proposed to avoid retraining the generator at every time-step on previous classes by applying EWC \cite{EWC} in the generative network. We pursue a similar goal with the key difference of utilizing a hard attention mechanism similar to the one described by \cite{HAT, mallya2018piggyback,mancini2018adding}. All three approaches make use of the techniques originally proposed in the context of binary-valued networks \cite{courbariaux2015training}. Herein, binary weights are specifically learned from a real-valued embedding matrix that is passed through a binarization function. To this end, \cite{mallya2018piggyback,mancini2018adding} learn to mask a pre-trained network without changing the weights of the base networks, whereas \cite{HAT} (HAT) features binary mask-learning for the layer activations simultaneously to the training of the base network. While DGMa features HAT-like layer activation masking, DGMw accomplishes binary mask learning directly on the weights of the generator. %
Other works propose to use non-binary filters to define a new task solving network in terms of a linear combination of the parameters of a fixed base network \cite{rosenfeld2018incremental, rebuffi2018efficient}. %

Similarly to \cite{yoon2018lifelong}, we propose to expand the capacity of the employed base network, in our case the samples generator. The expansion is performed dynamically with an increasing amount of attained knowledge. However, \cite{yoon2018lifelong} propose to keep track of the \emph{semantic drift} in every neuron, and then expand the network by duplicating neurons that are subject to sharp changes. In contrast, we compute weights’ importance concurrently during the course of network training by modeling the neuron behavior using learnable binary masks. As a result, our method explicitly does not require any further network retraining after expansion.

Other approaches like \cite{FearNet, kamra2017deep,schwarz2018progress} try to explicitly model short and long term memory with separate networks. %
In contrast to these methods, our approach does not explicitly keep two separate memory locations, but rather incorporates it implicitly in a single memory network. Thus, the memory transfer occurs during the binary mask learning from non-binary (short term) to completely binary (long term) values.

\section{Dynamic Generative Memory}
\label{sec:method}

Adopting the notation of \cite{RWalk}, let  $S_{t}=\{(x_{i}^{t}, \textit{y}_{i}^{t} )\}_{i=1}^{n^{t}} $ denote a collection of data belonging to the task $t \in T$, where $x_i^t \in \mathcal{X}$ is the input data and $\textit{y}_i^t \in \boldmath{\textit{y}^t}$ are the ground truth labels. While in the non-incremental setup the entire dataset $S =\cup_{t=0}^{|T|}S_{t}$ is available at once, in an incremental setup %
it becomes available to the model in chunks $S_t$ specifically only during the learning of task $t$.  Thereby, $S_{t}$ can be composed of a collection of items from different classes, or even from a single class only. %
Furthermore, at the test time the output space covers all the labels observed so far featuring the single head evaluation: %
$ \mathcal{Y}^t = \cup_{j=1}^{t}\textit{y}^j$.

We consider a continual learning setup, in which a task solving model $D$ has to learn its parameters $\theta^D$ from the data $S_t$ being available at the learning time of task $t$. Task solver $D$ should be able to maintain good performance on all classes $\mathcal{Y}^t$ seen so far during the training. A conventional DNN, while being trained on $S_t$, would adapt its parameters in a way that exhibits good performance solely on the labels of the current task $y^t$, the previous tasks would be forgotten. %
To overcome this, we introduce a Generative Memory component $G$, who's task is to memorize previously seen data distributions. As visualized in Fig.~\ref{fig1}, samples of the previously seen classes are synthesized by $G$ and replayed to the task solver $D$ at each step of continual learning to maintain good performance on the entire $\mathcal{Y}^t$.  %
We train %
a generative adversarial network (GAN)\cite{goodfellow2013empirical} \emph{and} a sparse mask for the weights of its generator simultaneously. The learned masks model connection plasticity of neurons, thus avoiding overwriting of important units by restricting SGD updates to the parameter segments of $G$ that exhibit free capacity.

\subsection{Learning Binary Masks} We consider a generator network $G_{\theta^G}$ consisting of $L$ layers, and a discriminator network $D_{\theta^D}$. In our approach, $D_{\theta^D}$ serves as both: a discriminator for generated fake samples of the currently learned task ($L_{adv.}$) and as a classifier for the actual learning problem ($L_{cls}.$) following the AC-GAN \cite{odena2016conditional} architecture. 
The system has to continually learn $T$ tasks. During the SGD based training of task $t$, %
we learn a set of binary masks $M^t = [m_1^t, ..., m_L^t]$ for the weights of each layer. %
Output of a fully connected layer $l$ is obtained by combining the binary mask $m_l^t$ with the layer weights:

\begin{equation}
y_l^t= \sigma_{act}[(m^t_l \circ W_l)^\top x],\quad W_l \in \mathbb{R}^{n\times p},
\label{eq:y_l}
\end{equation}
for $\sigma_{act}$ being some activation function. $W_l$ is the weight matrix applied between layer $l$ and $l-1$, $\cdot \circ \cdot$ corresponds to the Hadamard product. In DGMw $m_l^t$ is shaped identically to $W_l$, whereas in case of DGMa the mask $m_l^t$ is shaped as $1 \times p$ and should be expanded to the size of $W_l$. %
Extension to more complex models such as e.g. CNNs is straightforward.%

A single binary mask for a layer $l$ and task $t$ is given by: \begin{equation}
    m_l^t = \sigma(s\cdot e_l^t),
\end{equation} 
where $e_l^t$ is a real-valued mask embeddings matrix, $s$ is a positive scaling parameter $s \in \mathbb R_+$, and $\sigma$ a thresholding function $\sigma: \mathbb R \to [0,1]$. Similarly to \cite{HAT} we use the sigmoid function as a pseudo step-function to ensure gradient flow to the embeddings $e$. %
In training of DGMw, we anneal the scaling parameter $s$ incrementally during epoch $i$ from $1/s_{max}^i$ to $s_{max}^{i}$ (local annealing). $s_{max}^{i}$ is similarly adjusted over the course of $I$ epochs from $1/s_{max}$ to $s_{max}$ (global annealing with $s_{max}$ being a fixed meta-parameter). The annealing the scheme is largely adopted from \cite{HAT}:
\begin{align}
s_{max}^{i} &= \frac{1}{s_{max}} + (s_{max} - \frac{1}{s_{max}}) \frac{i-1}{I-1}\\
s &= \frac{1}{s_{max}^{i}} + (s_{max}^{i} - \frac{1}{s_{max}^{i}}) \frac{b-1}{B-1}.
\label{eq:mask_annealing}
\end{align}
Here $b \in \{1,\ldots,B\}$ is the batch index and $B$ the number of batches in each epoch of SGD training. DGMa only features global annealing of $s$, as it showed better performance.

In order to prevent the  overwriting of the knowledge related to previous classes in the generator network, gradients $g_l$ w.r.t. the weights of each layer $l$ are multiplied by the reverse of the cumulated mask $m_l^{\leq t}$: %
\begin{gather}
    g_l^{\prime}= [1 - m_{l}^{\leq t}] g_l, \quad   m_l^{\leq t} = \max(m_l^t, m_{l}^{t-1}),
\end{gather}
where $g_l^{\prime}$ corresponds to the new gradient matrix and $m_{l}^{\leq t}$ is the cumulated mask. %

Analogously to \cite{HAT}, we promote sparsity of the binary mask by adding a regularization term $R^t$ to the loss function of the AC-GAN\cite{odena2016conditional} generator:
\begin{gather}
    R^t(M^t, M^{t-1}) = \frac{\sum_{l=1}^{L-1} \sum_{i=1}^{N_l} m_{l,i}^{t}(1 - m_{l,i}^{<t}) }{ \sum_{l=1}^{L-1} \sum_{i=1}^{N_l} 1 - m_{l,i}^{<t} },
\label{eq:reg}
\end{gather}
where $N_l$ is the number of parameters of layer $l$. Here, parameters that were reserved previously are not penalized, promoting reuse of units over reserving new ones. %

\begin{table*}                            
    \newcommand{\rot}[1]{\rotatebox{90}{#1}}
    \newcommand{\multirot}[1]{\multirow{5}{*}{\rot{#1}}}
      \centering
    \begin{tabular}[b]{clcccccccccccc}
    
        \toprule                    
          &&&\multicolumn{2}{c}{MNIST (\%)}& &\multicolumn{2}{c}{SVHN(\%)} & &      \multicolumn{2}{c}{CIFAR10(\%)} & & \multicolumn{2}{c}{ImageNet-50(\%)}\\
         &Method   &  &  $A_{5} $ & $A_{10}$&&$A_{5}$ & $A_{10}$ & & $A_{5}$ & $A_{10}$ & & $A_{30}$ & $A_{50}$ \\
        \midrule    
         &JT  && 99.87  & 99.24 && 92.99 & 88.72 && 83.40 & 77.82 && 57.35 & 49.88\\\midrule  
        \multirow{4}{*}[0.5ex]{\rotcell{Episodic memory}}&iCarl-S \cite{iCarl} && - & 55.8 &&-&-&&-&-&& 29.38 & \textbf{28.98} \\
        &EWC-S\cite{EWC} && - &79.7       &&-&-&&-&-&&-&-\\
        &RWalk-S\cite{RWalk} && - & 82.5 &&-&-&&-&-&&-&-\\      
        &PI-S \cite{ZenkePG17}&& - & 78.7 &&-&-&&-&-&&-&-\\
        \midrule

        \multirow{5}{*}[0.5ex]{\rotcell{Generat. memory}} & EWC-M* \cite{seff2017continual}    && 70.62 & 77.03 && 39.84 & 33.02 &&-&-&&-&- \\
        &DGR* \cite{DGR}    && 90.39 & 85.40 && 61.29 & 47.28 &&-&-&&-&-\\
        &MeRGAN \cite{2018arXiv180902058W}    && 99.15 & 96.83 && 74.83 & 49.43 &&-&-&&-&-\\ \cmidrule{2-14}%
        
        &DGMw (ours)    && 98.75  & 96.46 && \textbf{83.93} & \textbf{74.38} &&\textbf{72.45}&\textbf{56.21}&& \textbf{32.14} & 17.82 \\           
        &DGMa (ours)  &&   \textbf{99.17} & \textbf{97.92} && 81.07 & 66.89 && 71.91 &51.75 &&25.93& 15.16 \\\hline
        \\
      \end{tabular}                  
    \caption{Comparison to the benchmark presented by \cite{RWalk} (episodic memory with real samples) and \cite{2018arXiv180902058W} (generative memory) of approaches evaluated in class-incremental setup. Joint training (JT) represents the upper bound (* aka. direct accuracy - classifier trained on real and tested on generated data [31]). 
    }
    \label{table_1}
\end{table*}

\label{sec:dynamicNetworkExpansion}        
\subsection{Dynamic Network Expansion}  As discussed by \cite{yoon2018lifelong}, significant domain shift between tasks leads to rapid network capacity exhaustion, manifesting in decreasing expressive power of the underlying network and ultimately in catastrophic forgetting. In case of DGM this effect will be caused by decreasing number of ``free ''  parameters over the course of training due to parameter reservation. %
To avoid this effect, we take measures to ensure constant number of free parameters for each task.  %

\textit{DGMa.} Consider a network layer $l$ with an input vector of size $n$, an output vector of size $p$, and the mask $m_l^1 \in [0,1]^{1 \times p}$ initialized with mask elements $m_l^1$ of all neurons of the layer set to 0.5 (real-valued embeddings $e_l^1$ are initialized with 0). After the initial training cycle on task $1$, the number of free output neurons in layer $l$ will decrease to $p - \delta_1$, where $\delta_t$ is the number of neurons reserved for a generation task $t$, here $t=1$. After the training cycle, the number of output neurons $p$ of the layer $l$ will be expanded by $ \delta_1 $. This guarantees that the free capacity of the layer is kept constant at $p$ neurons for each learning cycle.

\textit{DGMw.} In case of DGMw, after the initial training cycle %
the number of free weights will decrease to $np - \delta^{'}_1$, %
with $\delta^{'}_1$ corresponding to the number of weights reserved for the generation task $1$. %
The number of output neurons $p$ is expanded by  $\delta^{'}_1/n $. The number of free weights of the layer is kept constant, which can be verified by the following equation: $(p+\delta_t/n)n - \delta_t = np$.
In practice we extend the number of output neurons by $\lceil \delta_t/n \rceil$. %
The number of free weight parameters in layer $l$ is thus either $np$, if $\delta_t/n \in \mathbb{Z}$, or $np+p$, otherwise. %

\subsection{Training of DGM} The proposed system combines the joint learning of three tasks: a generative, a discriminative and finally, a classification task in the strictly class-incremental setup.

Using task labels as conditions, the generator network must learn from a training set $X_t = \{ X^t_1, ... , X^t_N \}$ to generate images for task $t$. To this end, AC-GAN's conditional generator synthesizes images $x_t = G_{\theta^G}(t, z, M_t)$, where $\theta^G$ represents the parameters of the generator network, $z$ denotes a random noise vector. %
The parameters corresponding to each task %
are optimized in an alternating fashion. As such, the generator optimization problem can be seen as minimizing $\mathcal{L_G} =\mathcal{L}_s^t - \mathcal{L}_c^t + \lambda_{RU} R^t$,  with $\mathcal{L}_c$ a cross entropy classification loss calculated on the the auxiliary output, $\mathcal{L}_s$ a discriminative loss function used on the adversarial output layer of the network (implemented to be compliant with architectural requirements of WGAN\cite{gulrajani2017improved}) %
, and $R^t$ the regularizer term expanded upon in equation \ref{eq:reg}. To promote efficient parameter utilization, taking into consideration the proportion of the network already in use, the regularization weight $\lambda_{RU}$ is multiplied by the ratio $\alpha = \frac{S_t}{S_{free}}$, where $S_t$ is the size of the network before training on task $t$, and $S_{free}$ is the number of free neurons. This ensures that less parameters are reused during early stages of training, and more during the later stages when the model already has gained a certain level of maturity.%

The discriminator is optimized similarly through minimizing $\mathcal{L_D} = \mathcal{L}_c^t + \mathcal{L}_s^t + \lambda_{GP} \mathcal{L}_{gp}^t $, where $\mathcal{L}_{gp}^t$ represents a gradient penalty term implemented as in \cite{gulrajani2017improved} to ensure a more stable training process. %

\section{Experimental Results}

We perform experiments measuring the classification accuracy of our system in a strictly class-incremental setup on the following benchmark datasets: MNIST \cite{lecun1998mnist}, SVHN \cite{netzer2011reading},  CIFAR-10 \cite{krizhevsky2014cifar}, and ImageNet-50 \cite{russakovsky2015imagenet}. Similarly to \cite{iCarl,2018arXiv180902058W, RWalk} we report an average accuracy (\textbf{$A_t$}) over the held-out test sets of classes $0 ... t$ seen so far during the training.

\textbf{Datasets.}   			     
The MNIST and SVHN datasets are composed of 60000 and 99289 images respectively, containing digits. The main difference is in the complexity and variance of the data used. SVHN's images are cropped photos containing house numbers and as such present varying viewpoints, illuminations, etc. CIFAR10 contains 60000 labeled images, split into 10 classes, roughly 6k images per class. Finally, we use a subset of the iILSVRC-2012 dataset containing 50 classes with 1300 images per category. All images are further resized to 32 x 32 before use.

\textbf{Implementation details.} We make use of the same architecture for the MNIST and SVHN experiments, a 3-layer DCGAN \cite{radford2015unsupervised}, with the generator's number of parameters modified to be proportionally smaller than in \cite{2018arXiv180902058W} (approx. $10\%$ of the DCGAN's generator size used by \cite{2018arXiv180902058W} for DGMw, and $44\%$ for DGMa on MNIST and SVHN). The projection and reshape operation is performed with a convolutional layer instead of a fully connected one.    
For the CIFAR-10 experiments, we use the ResNet architecture proposed by~\cite{gulrajani2017improved}. For the ImageNet-50 benchmark, the discriminator features a ResNet-18 architecture. All are modified to function as an AC-GAN\cite{odena2016conditional}.

All datasets are used to train a classification network in an incremental way. The performance of our method is evaluated quantitatively through comparison with benchmark methods. Note that we compare our method mainly to the approaches that rely on the idea of generative memory replay, e.g. replaying generator synthesized samples of previous classes to the task solver without storing real samples of old data. For the sake of fairness, we only consider benchmarks evaluated in class-incremental single-head evaluation setup. Hereby, to best of our knowledge \cite{2018arXiv180902058W} represent the state-of-the-art benchmark followed by \cite{DGR} and \cite{seff2017continual}. Next, we relax the strict incremental setup and allow partial storage of real samples of previous classes. Here we compare to the iCarl \cite{iCarl}, which is the state-of-the-art method for continual learning with storing real samples. 
\begin{figure}%
  \includegraphics[width=1\linewidth]{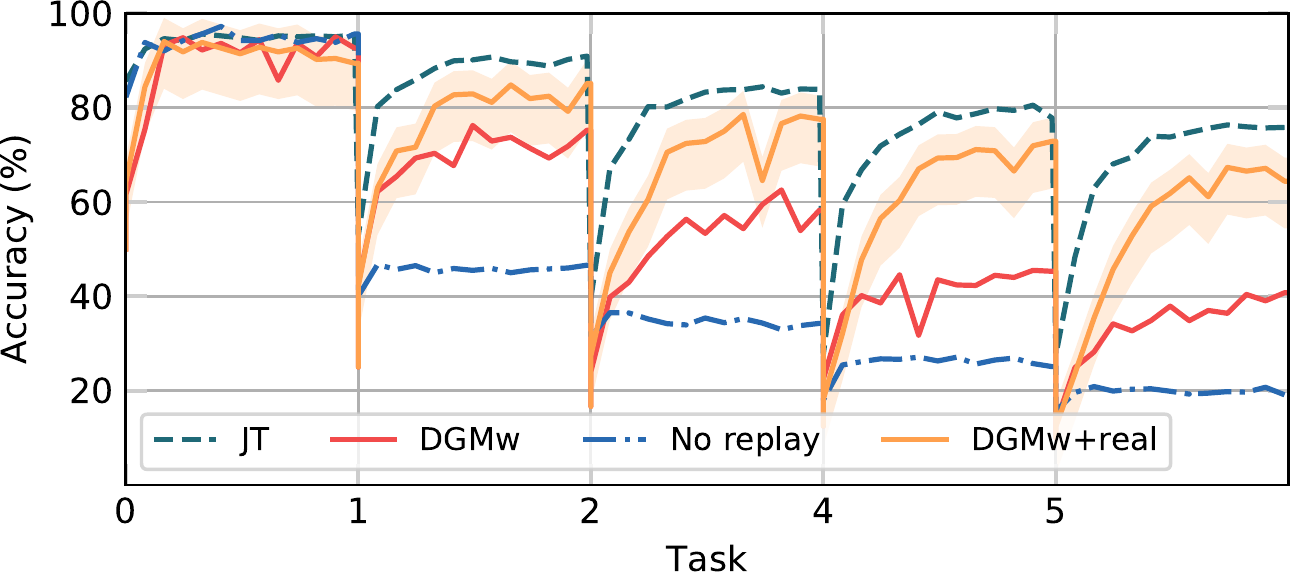}
  \caption{Top-5 performance of DGMw together with upper and lower performance bounds measured for ImageNet-50 benchmark. DGM+real denotes variation with different   ratios of real samples added to the replay loop (25\%-75\% of samples being real)}%
  \label{fig:ImgNetAcc}
 \end{figure}
  \begin{figure*}%
    \centering                
    \subfloat[Parameter re-usability ratios.]{{\includegraphics[width=0.48\linewidth]{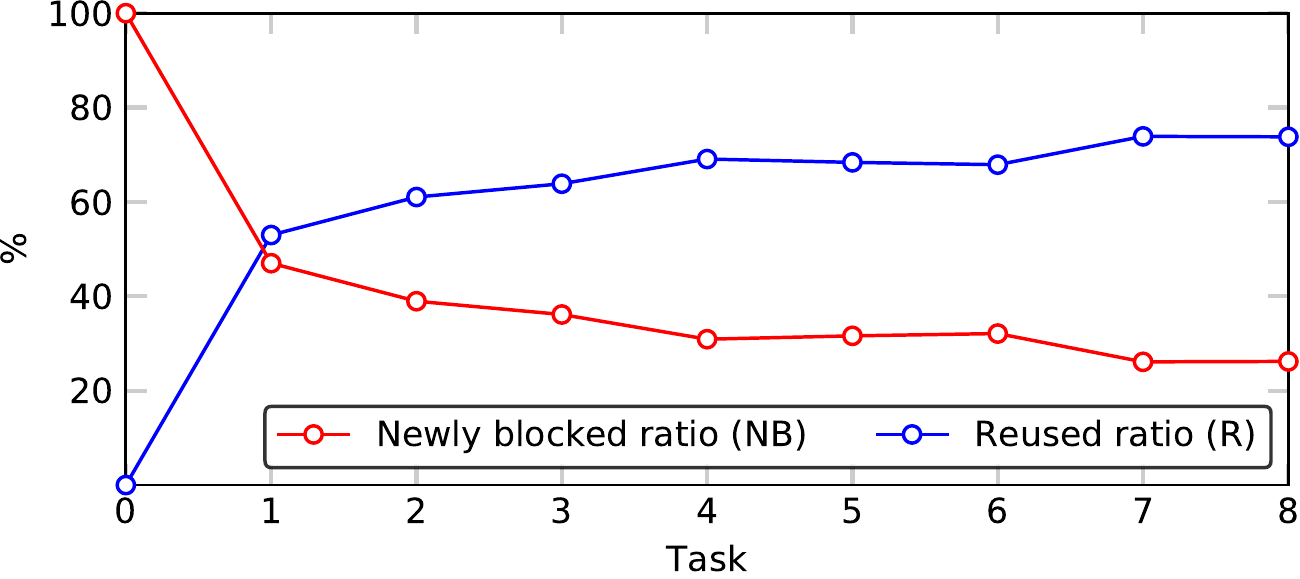}}} \hspace{0.5cm}
    \subfloat[Mask learning trajectories DGMa]{{\includegraphics[width=0.48\linewidth]{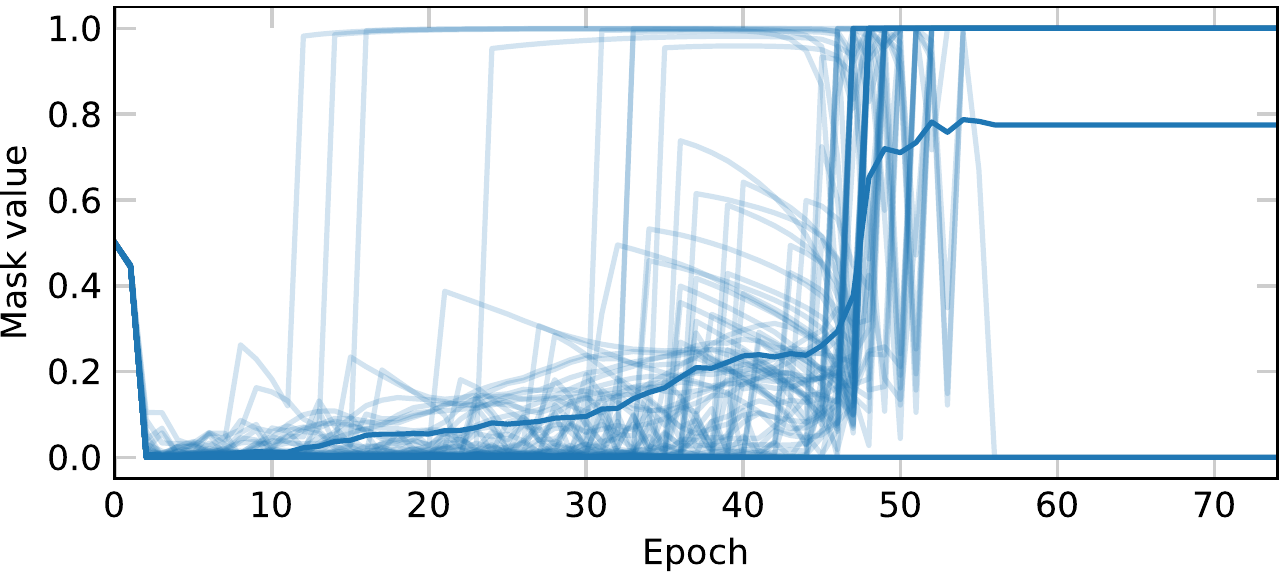}}}
    \caption{Masks learning dynamics. Fig. (a) illustrates the ratio of newly blocked and reused neurons over the total number of used neurons for a task $t$. Fig. (b) illustrates trajectories of mask value change for DGMa for a selected layer of G  (bold line - layer occupation).}%
    \label{fig:maskvals}%
\end{figure*}
\textbf{Results.} A quantitative comparison of both variants of the proposed approach with other methods is listed in Tab. \ref{table_1}. We use joint training (JT) as an upper performance bound, where the task solver $D$ is trained in a non-incremental fashion on all real samples without adversarial training being involved. The first set of methods evaluated by \cite{RWalk} do not adhere to the strictly incremental setup, 
 and thus make use of stored samples, which is often referred to as "episodic memory". %
 The second set of methods we compare with do not store any real data samples. %
 Our method outperforms the state of the art \cite{seff2017continual,DGR} on the MNIST and SVHN benchmarks through the integration of the memory learning mechanism directly into the generator, and the expansion of said network as it saturates to accommodate new information. We yield an increase in performance over \cite{2018arXiv180902058W}, a method that is based on a replay strategy for the generator and does not provide dynamic expansion mechanism of the memory network, leading to increased training time and sensitivity to semantic drift.  %
 As it can be observed for both, our method and \cite{2018arXiv180902058W}, the accuracy reported between 5 and 10-tasks of the MNIST benchmark has changed a little, suggesting that for this dataset and evaluation methodology both approaches have largely curbed the effects of catastrophic forgetting. DGM reached a comparable performance to JT on MNIST ($A_5$) using the same architecture. This suggests that the incremental training methodology forced the network to learn a generalization ability comparable to the one it would learn given all the real data. 
\begin{table}
      \small  
      \centering   
    \begin{tabular}[b]{lcccc}
        \toprule 
         & \multicolumn{2}{c}{Top-1(\%)} & \multicolumn{2}{c}{Top-5(\%)}\\
        Method & $A_{30}$ & $A_{50}$  & $A_{30}$ & $A_{50}$     \\
        \midrule
        
         JT  & 57.35 & 49.88 & 84.70 & 78.24 \\\hline 
        iCarl (K=1000) & 29.38 & 28.98 & 69.98 & 59.49          \\  
        iCarl (K=2000) & 39.38 & 29.96 & 70.57 & 60.07          \\\hline
        DGMw (K=1000) & 36.87 & 18.84 & 69.13 & 43.12 \\
        DGMw (K=2000) &  41.93 & 22.56 & 69.20 & 51.84     \\ \hline
        DGMw (r=0.75) & 50.80 & 38.22 & 78.27 & 64.64 \\
        DGMw (r=0.5)  & 48.80 & 40.72 & 79.40 & 71.72 \\                 
        DGMw (r=0.25) & 46.93 & 35.04 & 75.80 & 65.60     \\
        DGMw (r=0.1) & 41.67 & 30.48 & 71.80 & 61.04 \\
        DGMw (r=0)    & 32.14 & 17.82 & 62.53 & 40.76 \\ \bottomrule
      \end{tabular}
    \caption{Performance comparison of DGM and iCarl for different values of $r$ and memory size $K$.} %
    \label{tab:diff_r}
\end{table}
Given the high accuracy reached on the MNIST dataset largely gives rise to questions concerning saturation, we opted to perform a further evaluation on the more visually diverse SVHN dataset. In this context, increased data diversity translates to more difficult generation and susceptibility to catastrophic forgetting.  In fact, as can be seen in Tab.~\ref{table_1}, the difference between 5- and 10-task accuracies is significantly larger in all methods than what can be observed in the MNIST experiments. DGM strongly outperforms all other methods on the SVHN benchmark.  This can be attributed primarily to the efficient network expansion that allows for more redundancy in reserving representative neurons, and a less destructive joint use of neurons between tasks. Additionally, replay based methods (like \cite{DGR,2018arXiv180902058W}) can be potentially prone to generation of samples that represent class mixtures, especially for classes that semantically interfere with each other. DGM is immune to this problematic, since no generative replay is involved in the generator's training. Thus, DGM becomes more stable in the face of catastrophic forgetting. The quality of the generated images after 10 stages of incremental training for MNIST and SVHN can be observed in Fig.~\ref{fig:mnist_qual}. The generator is able to provide informative and diverse samples.

Finally, in the ImageNet-50 benchmark, we incrementally add 50 classes with 10 classes per step and evaluate the classification performance of DGM using single-head evaluation. The dynamics of the top-5 classification accuracy of our system is provided in Fig.~\ref{fig:ImgNetAcc}. 
Looking at the qualitative results shown in Fig.~\ref{fig:mnist_qual}, it can be observed that generated samples clearly feature class discriminative features which are not forgotten after incremental training on 5 tasks of the benchmark. 
Nevertheless, for each newly learned task the discriminator network's classification layer is extended with 10 new outputs, making the complexity of the classification problem to grow constantly (from 10-way classification to 50-way classification). With the more complex ImageNet samples also the generation task becomes much harder than in datasets like MNIST and SVHN. These factors negatively impact the classification performance of the task solver presented in Fig. \ref{fig:ImgNetAcc}, where DGMw performs significantly worse than the JT upper bound.   %

\begin{figure}[h!]%
    \centering         
    \includegraphics[width=\linewidth]{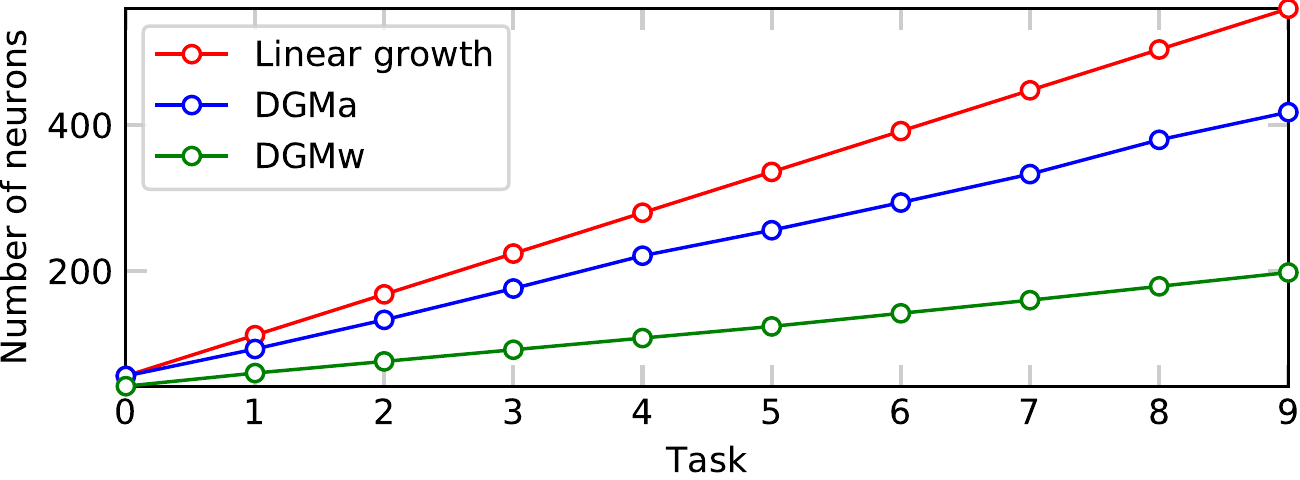}%
\caption{Network growth (numb. of neurons) in the incremental MNIST setup. Comparison of best performing DGMw and comparable DGMa (DGMw $A_{10}$: 96.46\%, DGMa - 96.98\%).%
} %
\label{fig:masksizes}%
\end{figure}

Next, we relax the strict incremental setup and allow the DGM to partially store real samples of previous classes. We compare the performance of DGM to the state-of-the-art iCarl \cite{iCarl}\footnote{ We use classes of ImageNet-50 with $32\times32$ resolution with the iCarl implementation under \url{https://github.com/srebuffi/iCaRL}}. Noteworthy, iCarl relies only on storing real samples of previous classes introducing a smart sample selection strategy. We define a ratio of stored real and total replayed samples $r=n_r/N$, where $N$ is the total number of samples replayed per class and $n_r$ is the number of randomly selected real samples stored per each previously seen class. To keep the number of replayed samples balanced with the number of real samples, $N$ is set to be equal to the average number of samples per class in the currently observed data chunk $S_t$. Furthermore, similarly to iCarl \cite{iCarl} we define $K$ to be the total number of real samples that can be stored by the algorithm at any point of time. We compare DGMw with iCarl for different values of $K$ %
allowing the storage of $K/|\mathcal{Y}^t|$ samples per class. %

From Tab.~\ref{tab:diff_r} we observe that DGM is outperformed by iCarl when no real samples are replayed (i.e. $r=0$) after 50 classes in top-1 and after 30 and 50 classes in top-5 accuracy. DGMw with $r=0$ outperforms iCarl with $K=1000$ in top-1 accuracy after 30 classes. Furthermore, we observe that adding real samples to the replay loop boosts DGM's classification accuracy beyond the iCarl's one. Thus, already for $r=0.1$ %
the performance of our system can be improved significantly. %
We now consider DGM and iCarl with the same memory size $K$ %
(we test for $K=1000$ and $K=2000$). Here DGM outperforms iCarl in top-1 accuracy after 30 classes, and almost reaches it in Top-5 accuracy. This is largely attributed to the advantage of DGM using generated samples additionally to the stored real once. Yet, a significant performance drop is observed after learning 5 tasks ($A_{50}$), where DGMw is outperformed by iCarl. This can be attributed to \textit{(a)} the fact that the number of samples replayed per class decreases over time due to fixed $K$ and increasing number of classes (e.g. for $K=2000$, 66 samples are played per class after seeing 30 classes, and 40 samples after 50 classes), as well as \textit{(b)} iCarl's smart samples selection strategy that favors samples that better approximate the mean of all training samples per class. Such samples selection strategy appears to works better in a situation where the number of real samples available per class decreases over time. It is noteworthy that iCarl's samples selection strategy can also be applied to DGM.  
\begin{figure*}[!t] 
\centering                                
    \includegraphics[width=1\linewidth]{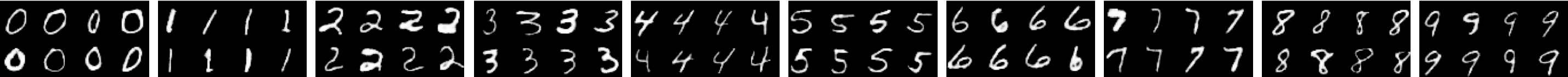}\\
    \centering 
    \includegraphics[width=1\linewidth]{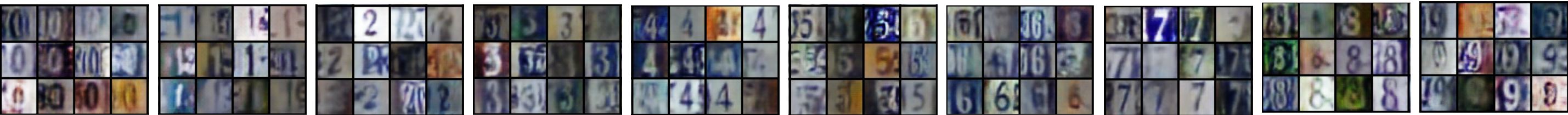}\\
    \centering 
    \hspace{0.001mm}
    \includegraphics[width=\linewidth]{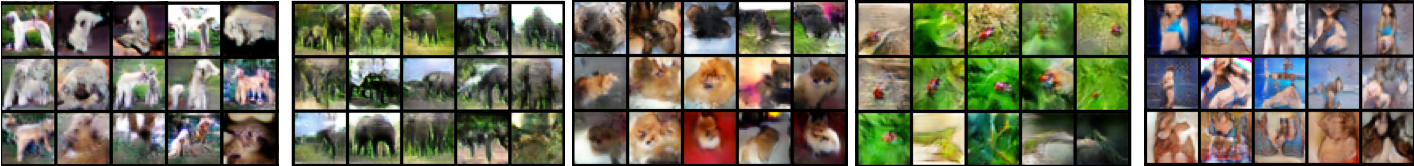}
    \caption{Images generated by DGM for MNIST(top), SVHN(middle) after training on 10 tasks, and ImageNet(bottom) after 5 tasks.} 
    \label{fig:mnist_qual}
\end{figure*}

\textbf{Growth Pattern Analysis.}
\label{sec:sive_vs_accuracy}      
One of the primary strengths of DGM is an efficient generator network expansion component, removing which would lead to the inability of the generator to accommodate for memorizing new task. %
Performance of DGM is directly related to how the network parameters are reserved during incremental learning,%
which ultimately depends on the generator's ability to generalize from previously learned tasks. Fig.~\ref{fig:masksizes} reports network growth against the number of tasks learned. We find that learning masks directly for the layer weights (DGMw) significantly slows down the network growth. Furthermore, one can observe the high efficiency of DGM's sub-linear growth  pattern as compared to the worst-case linear growth scenario. Interestingly, as shown in Tab.\ref{tab:size_comp}, after incrementally learning 10 classes the final number of generator's base network's parameters is lower than the one of the benchmarked MeRGAN \cite{2018arXiv180902058W}. More specifically, we observe the final network's size reduction of $26\%$ on MNIST, and $23\%$ on SVHN as compared to MeRGAN's fixed generator. In general, growth pattern of DGM depends on various factors: e.g. initialization size, similarity and order of classes etc.. A rather low saturation tendency of DGM's growth pattern observed in Fig.~\ref{fig:masksizes} can be attributed to the fact that with growing amount of information stored in the network, selecting relevant knowledge becomes increasingly hard. %

\begin{table}[]
    \centering                          \small
     \begin{tabular}[b]{ccccc} \toprule 
           Dataset & Method &   Size init. &  Size final & $A_{10}(\%)$   \\ \hline
         \multirow{2}{*}[0.5ex]{MNIST}& DGMw    &   $5.58\mathrm{e}{+4}$   & $3.83\mathrm{e}{+5}$ & 96.46 \\ 
         &MeRGAN  &   $5.18\mathrm{e}{+5}$   & $5.18\mathrm{e}{+5}$ & 97.00 \\\midrule
         \multirow{2}{*}[0.5ex]{SVHN} &DGMw    &   $5.58\mathrm{e}{+4}$   & $3.99\mathrm{e}{+5}$ & 74.38 \\ 
         &MeRGAN  &   $5.18\mathrm{e}{+5}$   & $5.18\mathrm{e}{+5}$ & 66.78  \\\hline
     \end{tabular}
     \caption{Comparison of generator's base network's size (number of parameters) of DGMw and MeRGAN.}
     \label{tab:size_comp}
 \end{table}

\textbf{Plasticity Evolution Analysis.}
We analyze how learning is accomplished within a given task $t$, and how this further affects the wider algorithm. For a given task $t$, its binary mask $M_t$ is initialized with the scaling parameter $s = 1$. Fig.~\ref{fig:maskvals}(b) shows the learning trajectories of the mask values over the learning time of task $t$. Here, at task initialization of DGMa the mask is completely non-binary (all mask values are 0.5). %
As training progresses, the scaling parameter $s$ is annealed, the network is encouraged to search for the most efficient parameter constellation (epoch 2-10). %
But with most mask values near 0 (most of the units are not used, high efficiency is reached), the network's capacity to learn is greatly curtailed. The optimization process pushes the mask to become less sparse, %
the number of non-zero mask values is steadily increasing until the optimal mask constellation is found, a trend observed in the segment between the epoch 10 and 55. %
This behaviour can be seen as a \emph{short-term memory} formation - if learning was stopped at e.g. epoch 40 only a relatively small fraction of learnable units would be masked in a binary way, the units with non-binary mask values would be still partially overwritten during the subsequent learning resulting in forgetting. A transition from short to the \emph{long-term memory} occurs largely within the epochs 45-65. Here the most representative units are selected and reserved by the network, parameters that have not made this transition are essentially left as unused for the learning task $t$. 
Finally, the optimal neuron constellation is optimized for the given task from epoch 60 onwards. %

For a given task $t$, masked units (neurons in DGMa, network weights in DGMw) can be broadly divided into three types: (i) units that are not used at all (U) [masked with 0] , (ii) units that are newly blocked for the task ($NB_t$), (iii) units that have been reused from previous tasks ($R_t$). Figure \ref{fig:maskvals}(a) presents the evolution of the ratio of the ($NB_t$) and ($R_t$) types over the total number of units blocked for the task $t$. Of particular importance is that the ratio of reused units is increasing between tasks, while the ratio of newly blocked units is decreasing. These trends can be justified by the network learning to generalize better, leading to a more efficient capacity allocation for new tasks.

\textbf{Memory Usage Analysis.}
We evaluate the viability of generative memory usage from the perspective of required disc space. Storing the generator for the ImageNet-50 benchmark (weights and masks) corresponds to the disc space requirement of $228 MB$. Thereby storing the preprocessed training samples of ImageNet-50 results in the required disc space of $315 MB$. In this particular case storing the generator $27.5\%$ more memory efficient than storing the training samples. Naturally, this effect will become more pronounced for larger datasets.

As discussed in Sec.~\ref{sec:sive_vs_accuracy}, DGMw features a more efficient network growth pattern as compared to DGMa. Yet, DGMw's attention masks are shaped identically to the weight matrices and thus require more memory. Tab. \ref{tab:memory_DGMa_DGMw} gives an overview of the required disc space for different components of DGMa and DGMw (masks are stored in a sparse form). Less total disc space is required to store DGMw's model as compared to DGMa, which suggests that DGMw's model growth efficiency compensates for the higher memory required for storing its masks. During the training, DGMw still exhibits a larger memory consumption, as the real-valued mask embeddings for the currently learned task must be kept in memory in a non-sparse form.

\begin{table}[!h]
    \small
    \centering    
     \begin{tabular}[b]{ccccc}  \toprule
                            &  \multicolumn{2}{c}{$Size_5$(MB)} &   \multicolumn{2}{c}{$Size_{10}$(MB)} \\ 
                            & DGMa & DGMw &  DGMa & DGMw  \\ \midrule
               Weights& 6.7     & 1.1 &  14.0  & 2.0    \\       
             Masks & 0.4 &  3.8    &  0.8 & 9.9   \\  \midrule
                Total   & 7.1 & \textbf{4.9} & 14.8 &  \textbf{11.9}   \\ \hline
     \end{tabular}
     \caption{Disc space required to store different components of DGMw and DGMa in Megabytes (MB) after 5 and 10 tasks. Compared models exhibit comparable performance on MNIST.}
     \label{tab:memory_DGMa_DGMw}
 \end{table}

\section{Conclusion}
In this work we study the continual learning problem in a single-head, strictly incremental context.
We propose a Dynamic Generative Memory approach for class-incremental continual learning. Our results suggest that DGM successfully overcomes catastrophic forgetting by making use of a conditional generative adversarial model where the generator is used as a memory module endowed with neural masking. %
We find that neural masking works more efficient when applied directly to layers' weights instead of activations. Future work will address the limitations of the DGM including missing  backward knowledge transfer and limited saturation of the network growth pattern. %

{\small
\bibliographystyle{ieee}
\bibliography{egpaper_final}
}

\newpage

\end{document}